# IdentiFace: A VGGNet-Based Multimodal Facial Biometric System

Mahmoud Rabea, Hanya Ahmed, Sohaila Mahmoud, Nourhan Sayed
Systems and Biomedical Department, Faculty of Engineering Cairo University

*Abstract*- The development of facial biometric systems has contributed greatly to the development of the computer vision field. Nowadays, there's always a need to develop a multimodal system that combines multiple biometric traits in an efficient yet meaningful way. In this paper, we introduce "IdentiFace" which is a multimodal facial biometric system that combines the core of facial recognition with some of the most important soft biometric traits such as gender, face shape and emotion. We also focused on developing the system using only VGG-16 inspired architecture with minor changes across different subsystems. This unification allows for simpler integration across modalities. It makes it easier to interpret the learned features between the tasks which gives a good indication about the decision-making process across the facial modalities and potential connection. For the recognition problem, we acquired a **99.2%** test accuracy for five classes with high intra-class variations using data collected from the FERET database[1]. We achieved **99.4%** on our dataset and **95.15%** on the public dataset[2] in the gender recognition problem. We were also able to achieve a testing accuracy of **88.03%** in the face-shape problem using the celebrity face-shape dataset [3]. Finally, we achieved a decent testing accuracy of **66.13%** in the emotion task which is considered a very acceptable accuracy compared to related work on the FER2013 dataset[4].

*Keywords*- Biometrics, Computer Vision, Deep Learning, Multimodal System, Facial Recognition

## Introduction

Think about the face as a map—filled with unique features like the shape of the eyes, the movement of the eyebrows, the curve of the lips, and other special details. We're going to talk about how this map helps us recognize people, understand their emotions, and even guess their gender. It's amazing how our faces hold so much information!

When we look at someone's face, we can do more than just recognize them. We can understand the emotions they're feeling—like happiness, sadness, or surprise—just by looking at their expressions. We can also make educated guesses about whether someone is a man or a woman based on their facial features. We can also predict a person's face shape. All of this is part of what we call facial biometrics.

Facial biometrics, in a nutshell, involves using these special facial features to identify, analyze emotions, and infer gender. But here's the kicker: for something to be considered a true biometric, it needs to meet the two-thirds rule; that means it must have a specific threshold of uniqueness—your face, for instance, has to be at least **66.67%** different from anyone else's for it to be a reliable biometric identifier.

And lucky for us, the face does possess this kind of accuracy which makes it one of the most leading noninvasive low-cost methods available for usage as a biometric, which is why many people opt for it, similar to us.

During our work, we adhered to this rule which played a pivotal role in guiding our approach towards facial biometrics. Understanding the significance of this benchmark, we conscientiously sought to ensure that any model or algorithm we developed met or exceeded this standard of distinctiveness, which led us to rigorously evaluate and refine our models. We conducted meticulous testing and analysis, measuring the uniqueness and accuracy of the facial features used for identification, emotion analysis, and gender inference. Our objective was clear: we wouldn't settle for any model that fell short of achieving a reliability threshold below the 68% mark.
We have also worked on collecting our own dataset, which we aimed to be as miscellaneous as possible, as an uncontrolled dataset, to help with generalizing our results.

## Related Work

### *2.1 Face Recognition*

The field of face recognition has witnessed significant advancements. Our study draws inspiration from the influential VGG-16 architecture proposed by Simonyan and Zisserman (2014)[5]. Compared to other conventional methods for facial recognition, Deep learning has been found to achieve more promising results [6] and that is what made us choose the VGG model for our system. Our model harnesses the depth and structure of VGG-16 to further refine and enhance the accuracy of face recognition systems.

### *2.2 Gender Classification*

Since it was first proposed in 2013 by Andrew Zisserman and Karen Simonyan[5], many researchers have tried to use VGGNet to perform gender classification on different datasets. Some papers showed that VGGNet-based gender classification can outcome existing architectures[7] while other papers tried to investigate challenging datasets and reach high accuracies [8]. Transfer learning using VGGNet has also shown promising performance on gender classification[9]. The main struggle when dealing with gender classification tasks based on a VGG arch is that you must have a large dataset to match the complexity of the model and to try to clean your data as much as you can.

### *2.3 Face-Shape Prediction*

Face-shape problems are considered a tricky task. The manual labeling of the data and how different shapes overlap each other make each model perform differently than the other. Many papers have addressed this problem indicating the trade-off between the high accuracy and the number of classes. The VGG arch, especially the pre-trained VGG-Face has been widely used to address such a problem [10]. The main limitation of this problem is the variations of poses and face alignment in the picture. This is typically addressed by applying 68-landmark related networks that detect the face shape from the connections between the landmarks making it prone to many pose variations.

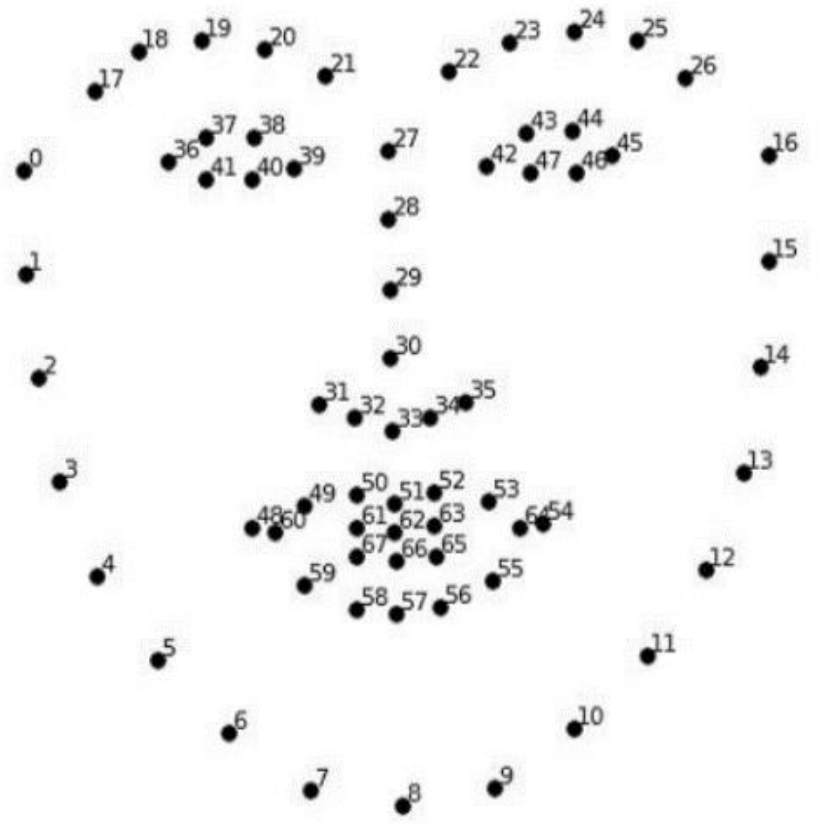

[1] Dlib 68-landmarks created by Brandon Amos of CMU who works on OpenFace.

### *2.4 Emotion Recognition*

Emotion recognition using face detection and normalization was proposed by Cui et al. in 2016. The CNN classifier was utilized as multiple classifiers for different face regions. If CNN is applied to the entire face image, then first frame the CNN for the mouth area, and subsequently for the eye area, as is likely done for other areas where separate CNNs are framed. The recognition accuracy of happy, sad, disgust and surprise expressions achieved no less than 98 %, while recognition accuracy of the angry and fear expressions was a little lower at about 96.7 % and 94.7 %, respectively. [11]

Zhang et al. used a different approach of localization then deploying the CNN architecture as well. [12]

Clawson et al. [13] observed that specific facial areas exhibit more prominent features for certain subtle emotional expressions. Leveraging this insight, they compare the accuracy of 'full-face' CNN modeling against upper and lower facial region models for recognizing emotions in static images. Additionally, they propose a human-centric CNN hierarchy achieved by histogram equalization and deploying a deep learning model. This hierarchy significantly boosts classification accuracy compared to separate CNN models, achieving a 93.3% true positive classification rate overall.

## Dataset

### *3.1 Face Recognition*

For the recognition task, we employed the Color FERET dataset [1] from NIST, containing 11,338 facial images of 994 individuals. This dataset encompasses 13 distinct poses, each annotated with the degree of facial rotation. Moreover, certain subjects have images with and without glasses, while others exhibit diverse hairstyles across their pictures. We specifically utilized the scaled-down versions of these images, sized at 256 x 484 pixels. This dataset was selected for its wide array of variations, aiding in training models to generalize effectively to new subjects. Additionally, we augmented the database by incorporating four new subjects, enabling us to test it across various scenarios, including ourselves in different variations.

### *3.2 Gender Classification*

For the gender problem, we collected our dataset from members of the faculty. The data initially consisted of 15 unique males and 8 unique females with most of them having more than one image with multiple variations to increase the data size. We then increased the number of unique subjects and ended up with 31 unique males and 27 unique females with a total number of images (133 males / 66 females). No Training/Validation data split was done during the collection process and it was done during the preprocessing phase.

For the sake of comparison, we chose a popular Gender Recognition dataset [2]. The dataset was split into training data with almost 23000 images per class and about 5500 images per class for validation. We chose this particular dataset as it has proved its efficiency for over 4 years, is well-preprocessed, and well-structured.

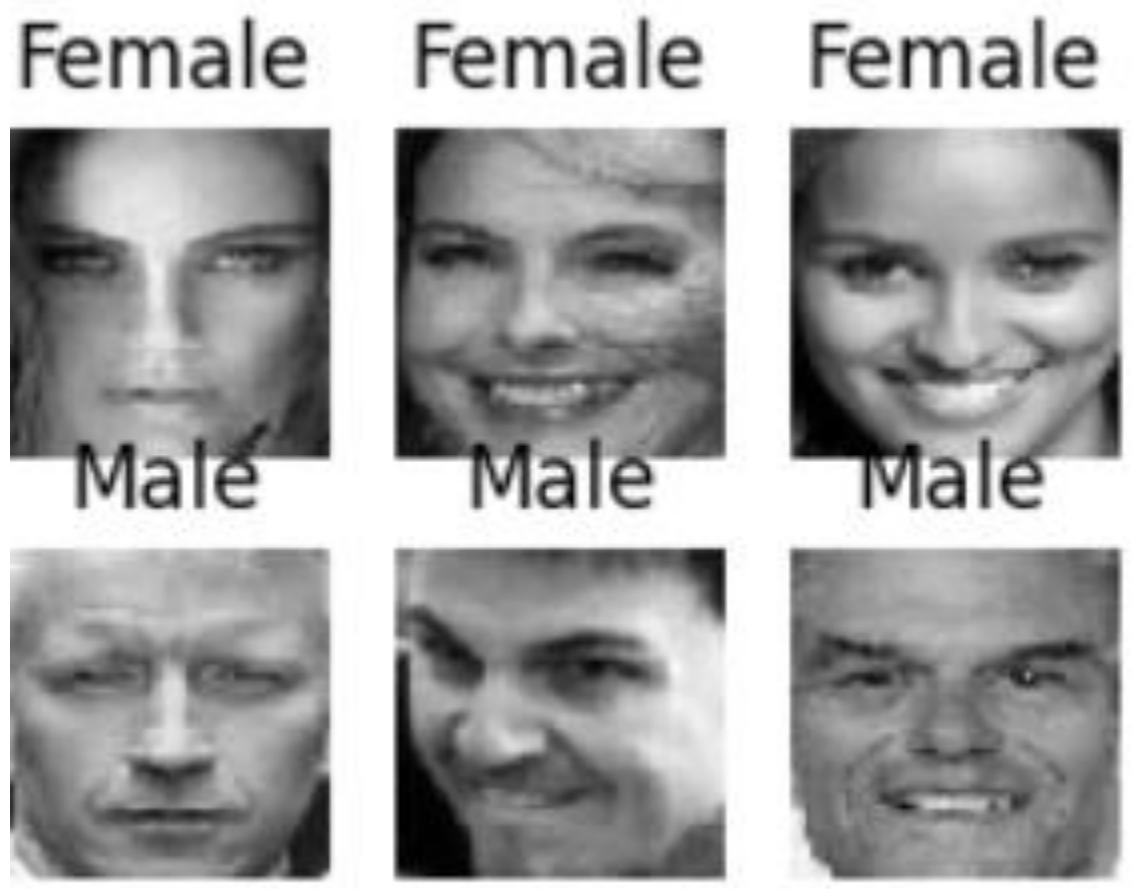

[2] Public Gender Dataset

### *3.3 Face-Shape Prediction*

Due to the complexity of this task. We couldn’t collect our dataset as it required manual labeling which isn’t a best practice. We chose to work with the most popular face shape data which is celebrity face shape [3]. The dataset was published back in 2019 and consisted of only female subjects with 100 images per class for a total of five classes (Round / Oval / Square / Oblong / Heart).

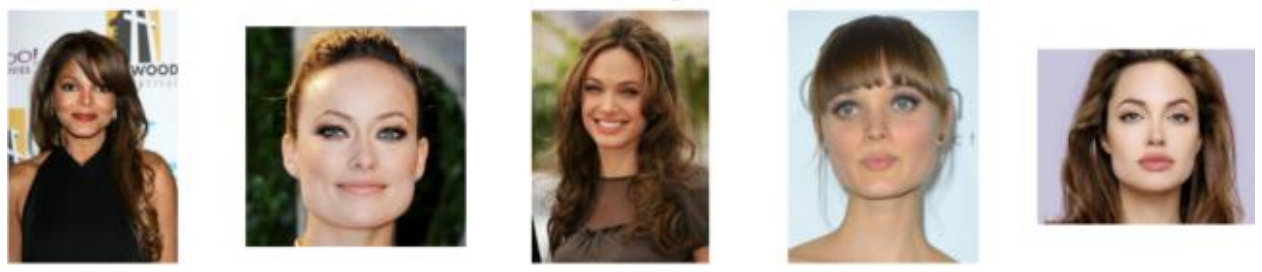

[3] Face-shape Dataset square samples

### *3.4 Emotion Recognition*

For this task, we first collected our dataset, which was 38 subjects divided into 22 males and 16 females. Each subject had a total of 7 images, each per a particular emotion, giving a total of 266 images with 38 images per class. Images of each class were labeled manually, which was challenging due to the variety between the subjects when asked to show a specific emotion. Also, some of them had similar facial expressions for more than one class, which made labeling the images and the classification process way more challenging as the classes were overlapping. To overcome the challenge of collecting a proper emotion dataset. We used the FER2013 dataset [4], which is public and consists of over 30000 images with 7 main classes: (Angry/Disgust/Fear/Happy/Sad/Surprise /Neutral). All images are converted into 48x48 grayscale images with an almost balanced distribution across all classes.

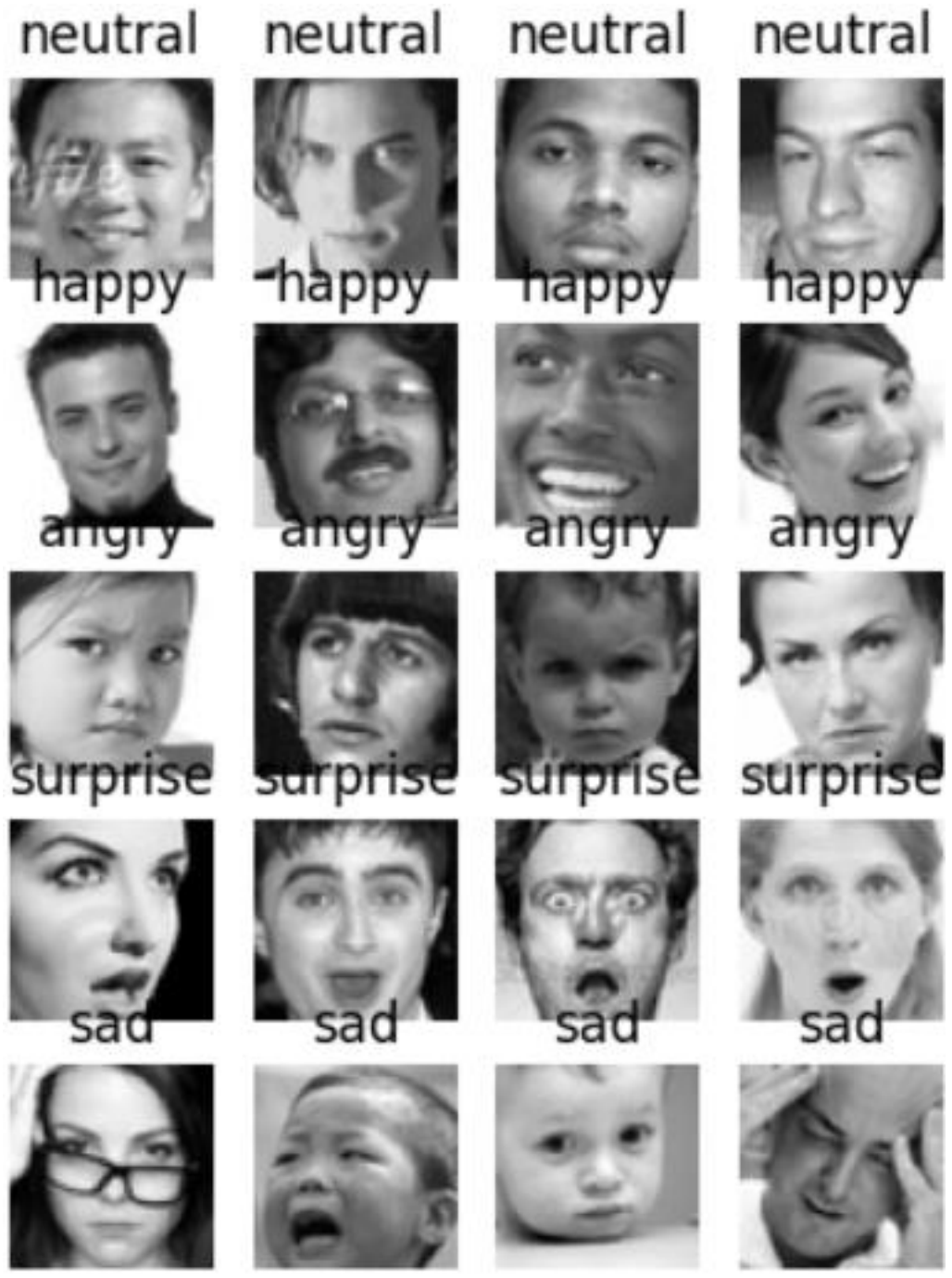

[4] FER2013 dataset

## Methodology

We aimed to have a single Network that can adapt to multiple facial problems with some minor changes between each problem. After experimenting with different architectures, we chose VGGNet architecture as our main network for the multimodal system.

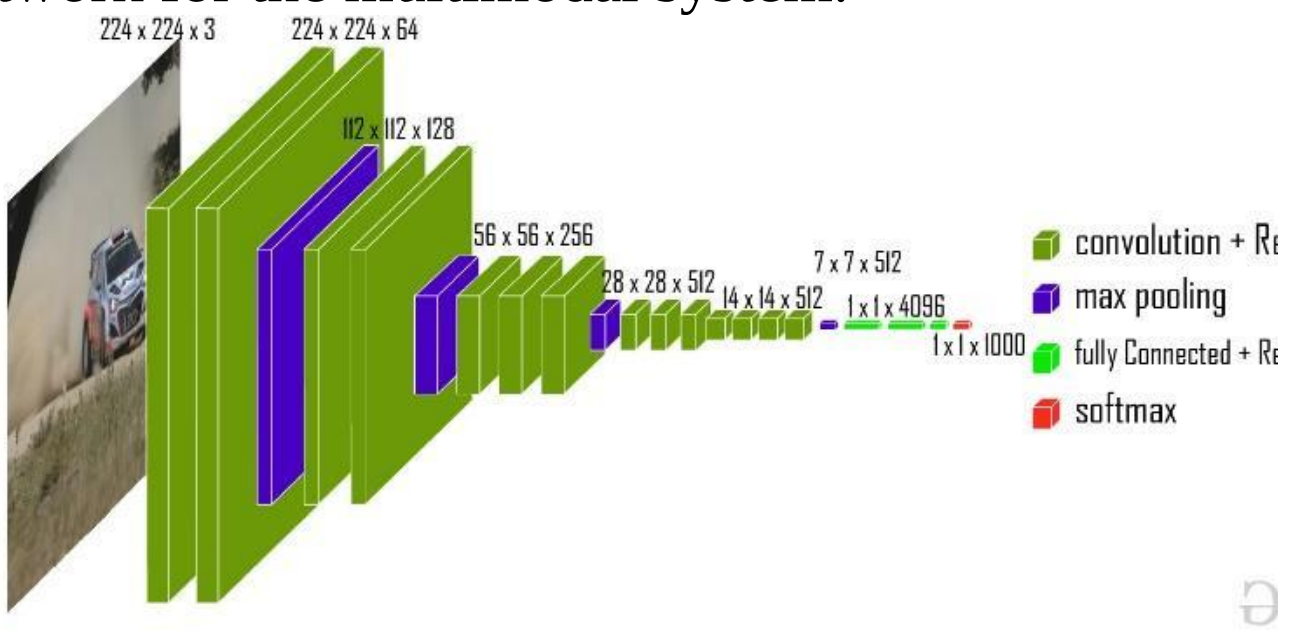

[5]"VGG-16: CNN model," GeeksforGeeks, https://www.geeksforgeeks.org/vgg-16-cnn-model/

We experimented with basic VGG-16, and we ended up simplifying it by only having 3 main blocks and removing the last 2 convolutional blocks. This was mainly done to reduce the number of parameters and the overall complexity of the model since it was already performing well on the various tasks. A general summary of the model with the number of layers, output shapes, and the number of parameters is provided in the following table:

| Layer | output shape | number of parameters |
|---|---|---|
| Conv2D | (None, 128, 128, 64) | 640 |
| Conv2D | (None, 128, 128, 64) | 36928 |
| MaxPooling2D | (None, 64, 64, 64) | 0 |
| Conv2D | (None, 64, 64, 128) | 73856 |
| Conv2D | (None, 64, 64, 128) | 147584 |
| MaxPooling2D | (None, 32, 32, 128) | 0 |
| Conv2D | (None, 32, 32, 256) | 295168 |
| Conv2D | (None, 32, 32, 256) | 590080 |
| Conv2D | (None, 32, 32, 256) | 590080 |
| MaxPooling2D | (None, 16, 16, 256) | 0 |
| Flatten | (None, 65536) | 0 |
| Dense | (None,512) in all tasks except for the emotion task (None,2048) | 33554944 |
| Dropout | 0.5 in all tasks | 0 |
| Dense / Classification layer | Depends on the task | |

Figure (1) Model Summary

Finally when compiling the model, We applied an Adam optimizer with sparse categorical cross entropy as our loss function. An Early Stopping is also present to prevent the model from overfitting.

### *4.1 Preprocessing*

For all the tasks but for face recognition, a general preprocessing is applied as follows:

1. A face detection method is applied using Dlib 68 facial landmarks.
2. all the detected faces are then cropped and the images with no faces are filtered.
3. the faces are then resized & grayscale (128,128)

Once the resizing is done, each dataset is augmented differently to ensure a balanced dataset across all tasks.

However, for the face recognition task, the preprocessing was as follows:

1. Face detection using Dlib's CNN-based face detection.
2. Cropping the identified faces and transforming them into grayscale images.
3. Resizing to 128x128 pixels
4. Changing the number of classes to five: Hanya, Mahmoud, Nourhan, Sohaila and Other.

Note that some of the public datasets were already preprocessed so we only performed a checking step to ensure the data is preprocessed the way we desire.

### *4.2 Augmentation*

Augmentation was done only to the unbalanced & small datasets to ensure a fair distribution across all classes. Different Augmentation techniques included:

| Technique | Variation | Applied to |
|---|---|---|
| Horizontal Flip | - | our gender dataset |
| | | Face recognition dataset |
| | | Face-shape dataset |
| Rotation | 30 left | our gender dataset |
| | 30 right | |
| | 15 left | |
| | 15 right | |
| | 10 left | Face-shape dataset |
| | 10 right | |
| | 5 left | |
| | 5 right | |
| | 10 right | Face recognition dataset |
| | 10 left | |

Figure (2) Augmentation Techniques

| Dataset | Total number of images per class before augmentation | Total number of images per class before augmentation | Augmentation factor for the single image |
|---|---|---|---|
| our gender dataset | Male: 133<br>Female: 66 | Male: 2500<br>Female: 2500 | Male: 107<br>Female: 221 |
| Face-shape dataset | Round: 93<br>Oval: 95<br>Square: 100<br>Oblong: 100<br>Heart: 99 | Round: 558<br>Oval: 570<br>Square: 600<br>Oblong: 600<br>Heart: 594 | Round: 6<br>Oval: 6<br>Square: 6<br>Oblong: 6<br>Heart: 6 |
| Face Recognition | Hanya: 55<br>Mahmoud: 100<br>Nour: 50<br>Sohaila: 34 | Hanya: 55<br>Mahmoud: 100<br>Nour: 50<br>Sohaila: 34 | Hanya: 9<br>Mahmoud: 5<br>Nour: 10<br>Sohaila: 14 |

Figure (3) Augmentation Results

For face recognition, we initially had 11,338 images for the “Other” class obtained from the color FERET dataset, and so we reduced it to 500 to avoid overfitting.
Some datasets like FER & the public Gender Recognition dataset didn’t need to be augmented as the distribution was balanced with many images per class.

## Results and Discussion

### *5.1 Face Recognition*

A train-test split ratio of 80-20 was used on our processed and augmented dataset. The following parameters were used to train our model:

- lr = 0.0001
- batch_size = 32
- test_size = 0.2
- epochs = 100

| Model | Train | | Test | |
|---|---|---|---|---|
| | Loss | Accuracy | Loss | Accuracy |
| | 0.0099 | 99.7% | 0.0322 | 99.2% |

Figure (4) Recognition evaluation

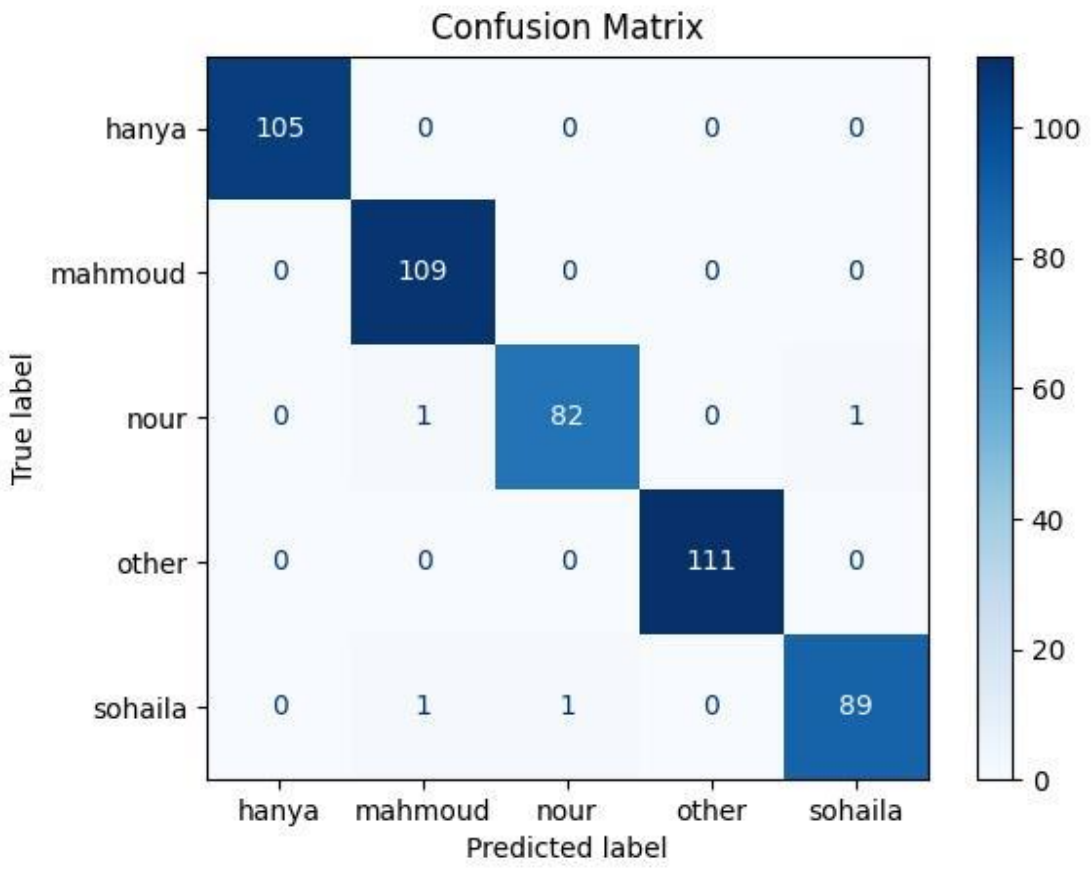

Figure(5) Recognition confusion matrix

*5.2 Gender Classification*

Instead of addressing this task as a binary task, we viewed the Gender classification problem as a multi-class classification problem labeling female subjects with 0 and male subjects with 1.

The following parameters were used to train both models (the public dataset model & our dataset model):

- lr = 0.0001
- batch_size = 128
- test_size = 0.2
- epochs = 3 & patience = 2 for our dataset while it was 15 & 3 for the public dataset respectively

| Model | Train | | Test | |
|---|---|---|---|---|
| | Loss | Accuracy | Loss | Accuracy |
| our dataset | 0.0412 | 99.5% | 0.0443 | 99.42% |
| Public dataset | 0.1027 | 96,48% | 0.1340 | 95.15% |

Figure (6) Gender Evaluation

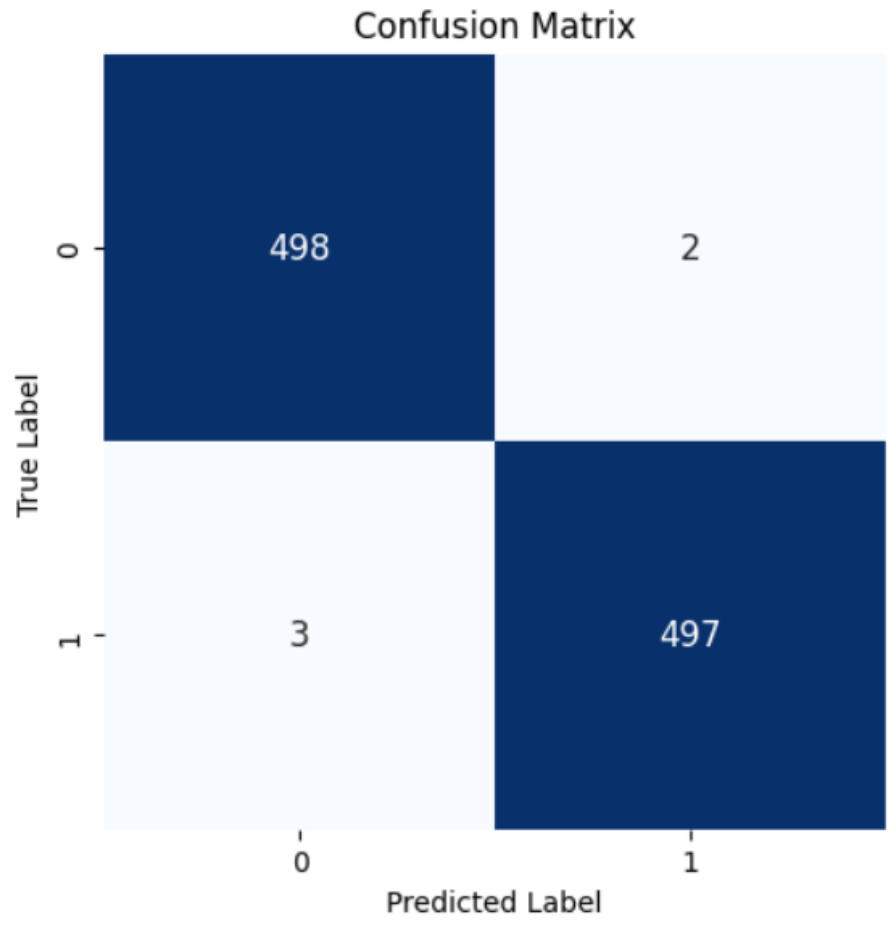

Figure (7) collected dataset confusion matrix

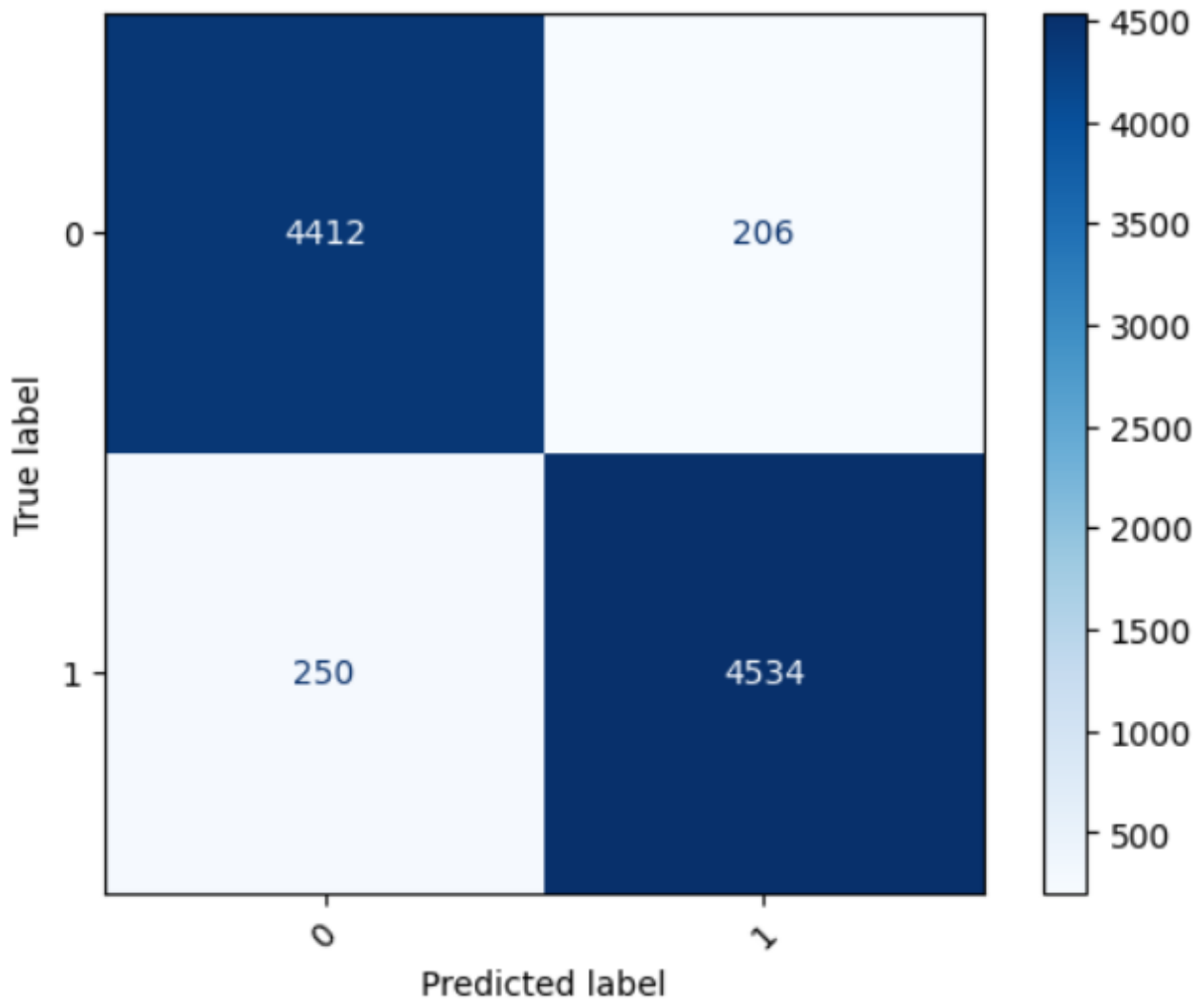

Figure (8) public dataset confusion matrix

| class | Precision | Recall | f1-score |
|---|---|---|---|
| Female | 95% | 96% | 95% |
| Male | 96% | 95% | 95% |

Figure (9) classification report for the final used model (public dataset model)

As observed, the two models achieved outstanding scores mainly due to the good quality data and the fact that the gender classification task is considered relatively easy compared to the other prediction tasks related to the face as biometric.

## *5.3 Face-Shape Prediction*

To address this task, we tried two different models, one for all classes and another model for only three classes (oblong/square / round). This was done to observe how the model will perform with classes that minimally overlap and compare it with the other model containing all classes.

| Model | oblong | square | round | heart | oval |
|---|---|---|---|---|---|
| 3 classes | 0 | 1 | 2 | - | - |
| All classes | 0 | 1 | 2 | 3 | 4 |

Figure (10) Face-shape labeling

For the two models, the following parameters were used:

- lr = 0.0001
- batch_size = 128
- test_size = 0.2
- epochs = 30 & patience = 7

| Model | Train | | Test | |
|---|---|---|---|---|
| | Loss | Accuracy | Loss | Accuracy |
| 3 classes | 0.0181 | 99.64% | 0.1942 | 94.03% |
| All classes | 0.0167 | 99.79% | 0.4485 | 88.03% |

Figure (11) Face-shape Evaluation

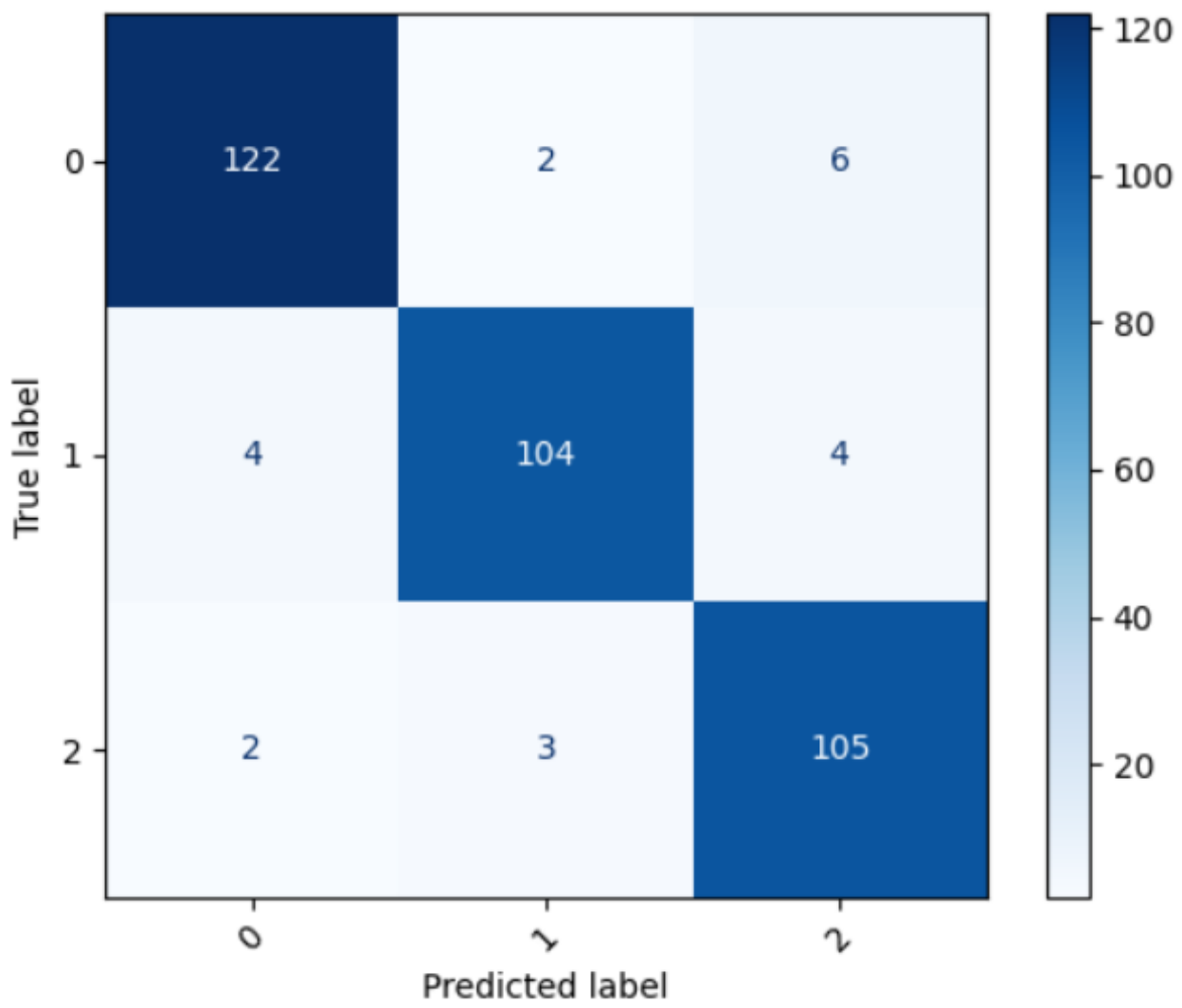

Figure (12) 3 classes confusion matrix

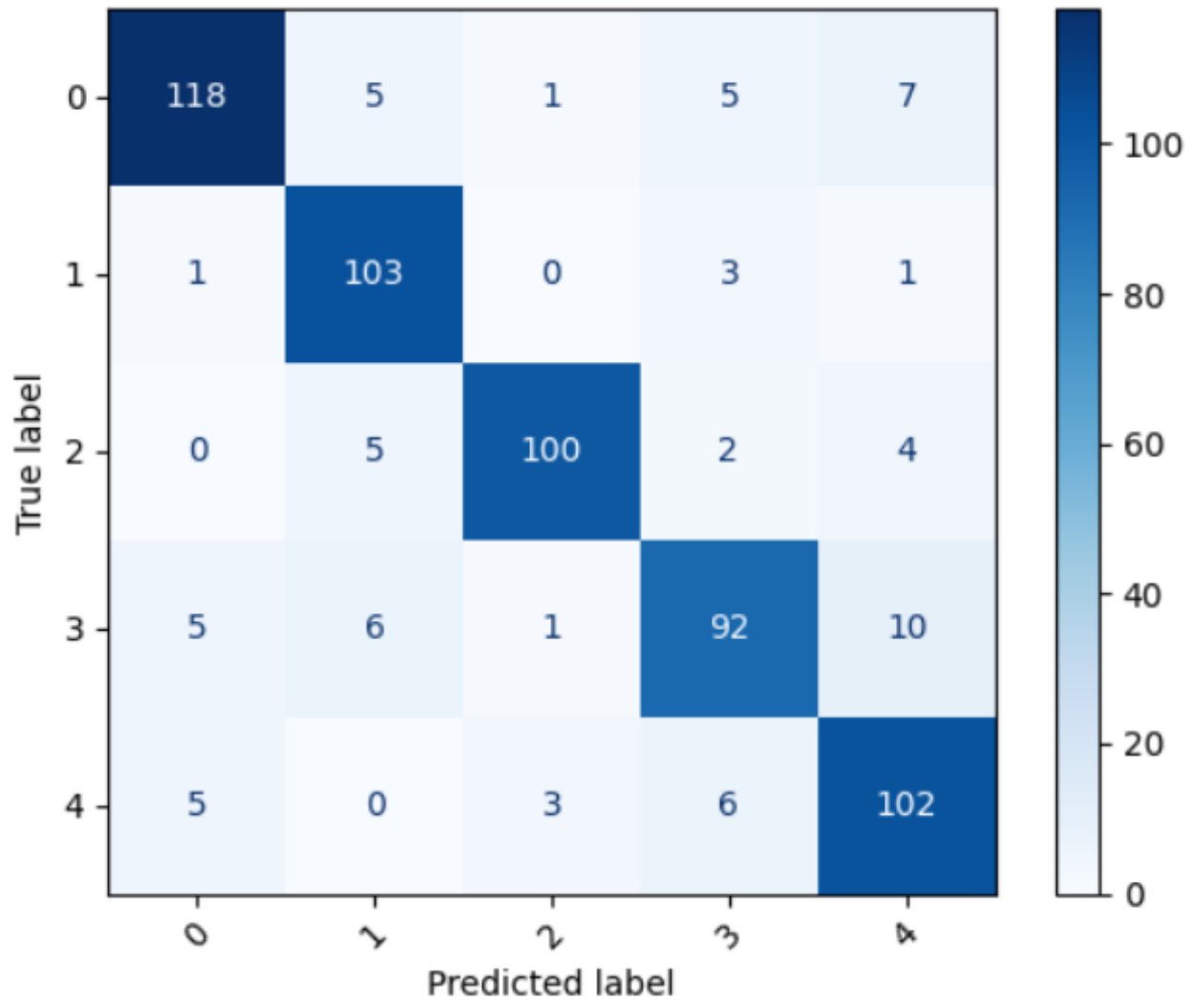

Figure (13) all classes confusion matrix

| class | precision | recall | f1-score |
|---|---|---|---|
| oblong | 91% | 87% | 89% |
| square | 87% | 95% | 91% |
| round | 95% | 90% | 93% |
| heart | 85% | 81% | 83% |
| oval | 82% | 88% | 85% |

Figure (14) classification report for the final used model (all classes model)

The provided results show that when increasing the number of classes and due to overlapping, the model starts to confuse similar classes. One thing we tried was to compare the prediction of our model with famous websites and the results were very subjective. Each website produced a different prediction that's mainly due to the data they used for training. Further improvements can be made to the current results by filtering the dataset or combining similar classes.

### *5.4 Emotion Recognition*

We divided our dataset manually into train and test with a split ratio of 70-30 to hide some subjects in the training data to ensure the models focus on learning emotional features and not on facial recognition. We used two approaches in this recognition problem. They are Support Vector Machine (SVM) and Convolutional Neural Networks (CNN).

#### *5.4.1 SVM*

We used various techniques for the features given to the SVM model. Also, due to the significant similarities between the classes, we tried to drop from 7 to 3 emotions to obtain minimum overlapping. The emotions are fear, anger, and happiness. The results of all used techniques are shown in Figure(15). The highest accuracy achieved using SVM was the 3-classes-68-landmarks SVM model, with an accuracy of 83%, and its confusion matrix is shown in Figure(16).

| Features extracted | Number of classes | Accuracy | Precision | Recall | F1 Score |
|---|---|---|---|---|---|
| face features | 7 | 24% | 23% | 24% | 23% |
| face features | 3 | 67% | 69% | 67% | 66% |
| 68 landmarks | 7 | 34% | 35% | 34% | 34% |
| 68 landmarks | 3 | 83% | 84% | 83% | 83% |
| LBP features | 3 | 47% | 48% | 47% | 42% |
| GF features | 3 | 30% | 19% | 30% | 20% |

Figure (15) SVM Models Results on our dataset

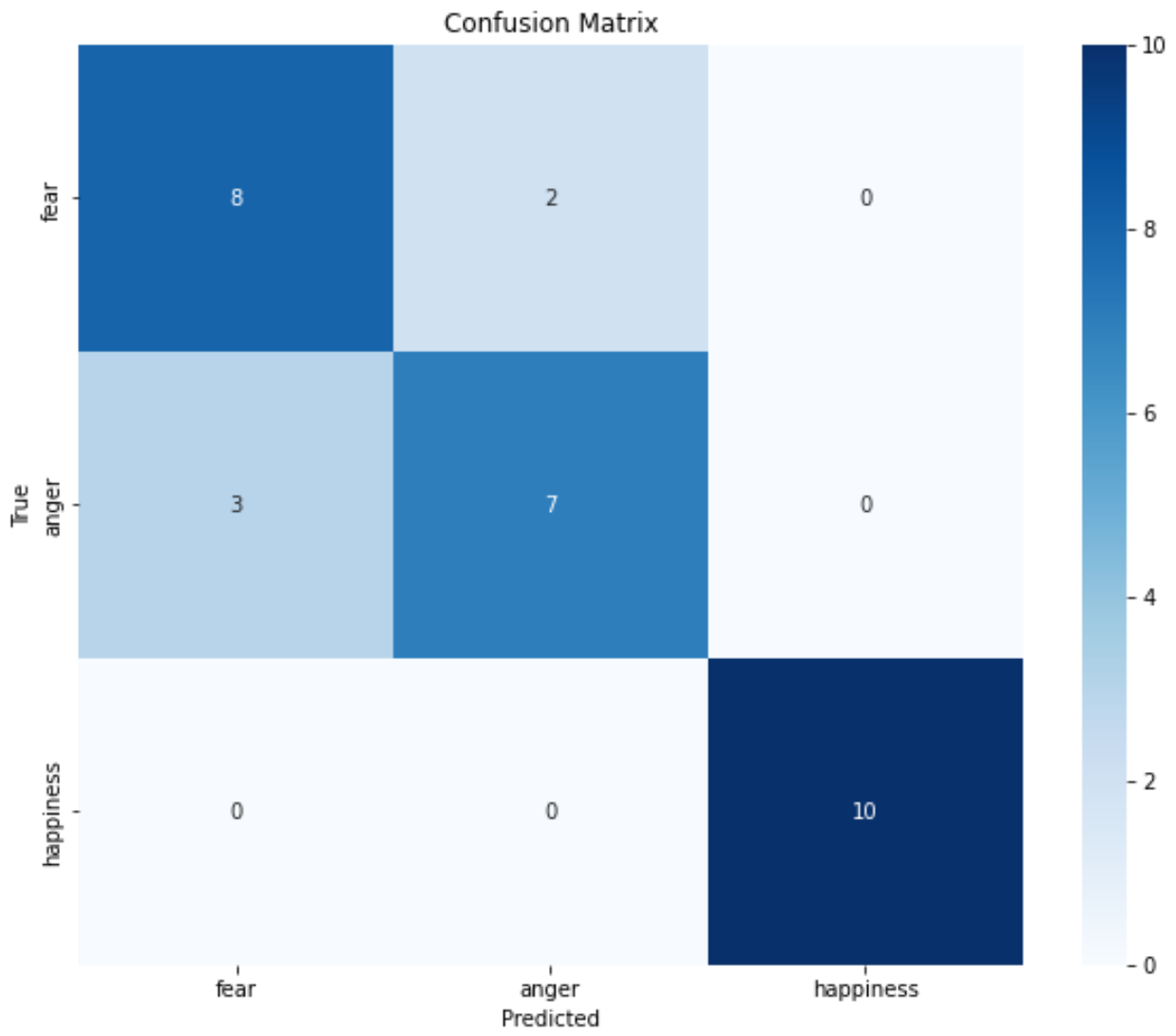

Figure (16) 3 emotions-68-landmarks-SVM confusion matrix

#### *5.4.2 CNN*

Firstly, we tried different CNN architectures, simple and complex, on our dataset and achieved lower accuracies than SVM as the dataset is too small, even after applying augmentation, to achieve high accuracies in a deep learning approach that requires a large dataset for good results. The results of some CNN models are shown in Figure(17).

| Model | Model Parameters | Train | | Test | |
|---|---|---|---|---|---|
| | | Loss | Accuracy | Loss | Accuracy |
| Base model with no dense layer | lr = 0.0001 epochs = 5 | 2.9015 | 19.89% | 2.7039 | 12.5% |
| simple model with one dense layer | lr = 0.0001 batch size = 32 epochs = 10 | 2.458 x10(-5) | 100% | 7.6586 | 19.02% |

Figure (17) Results of CNN models on our dataset

*5.4.3 VGG*

Lastly, we used the FER2013 dataset to enhance the results using VGGNet. The dataset is considered a complex challenge having an average test accuracy of 60-65%. To encounter this, we tried to filter the data from the 2 emotions with the most noise (Disgust and fear). We also tried at first a model without the sad emotion to address how this emotion would reflect on the behavior of the model.

| Model | neutral | happy | angry | surprise | sad |
|---|---|---|---|---|---|
| four emotions | 0 | 1 | 2 | 3 | - |
| five emotions | 0 | 1 | 2 | 3 | 4 |

Figure (18) Emotions labeling

For the two models, the following parameters were used:

- lr = 0.0001
- batch_size = 128
- test_size = 0.2
- epochs = 40 & patience = 7

| Model | Train | | Test | |
|---|---|---|---|---|
| | Loss | Accuracy | Loss | Accuracy |
| four emotions | 0.3920 | 86.94% | 0.7201 | 73.14% |
| five emotions | 0.5483 | 81.26% | 0.9161 | 66.13% |

Figure (19) Emotion Evaluation

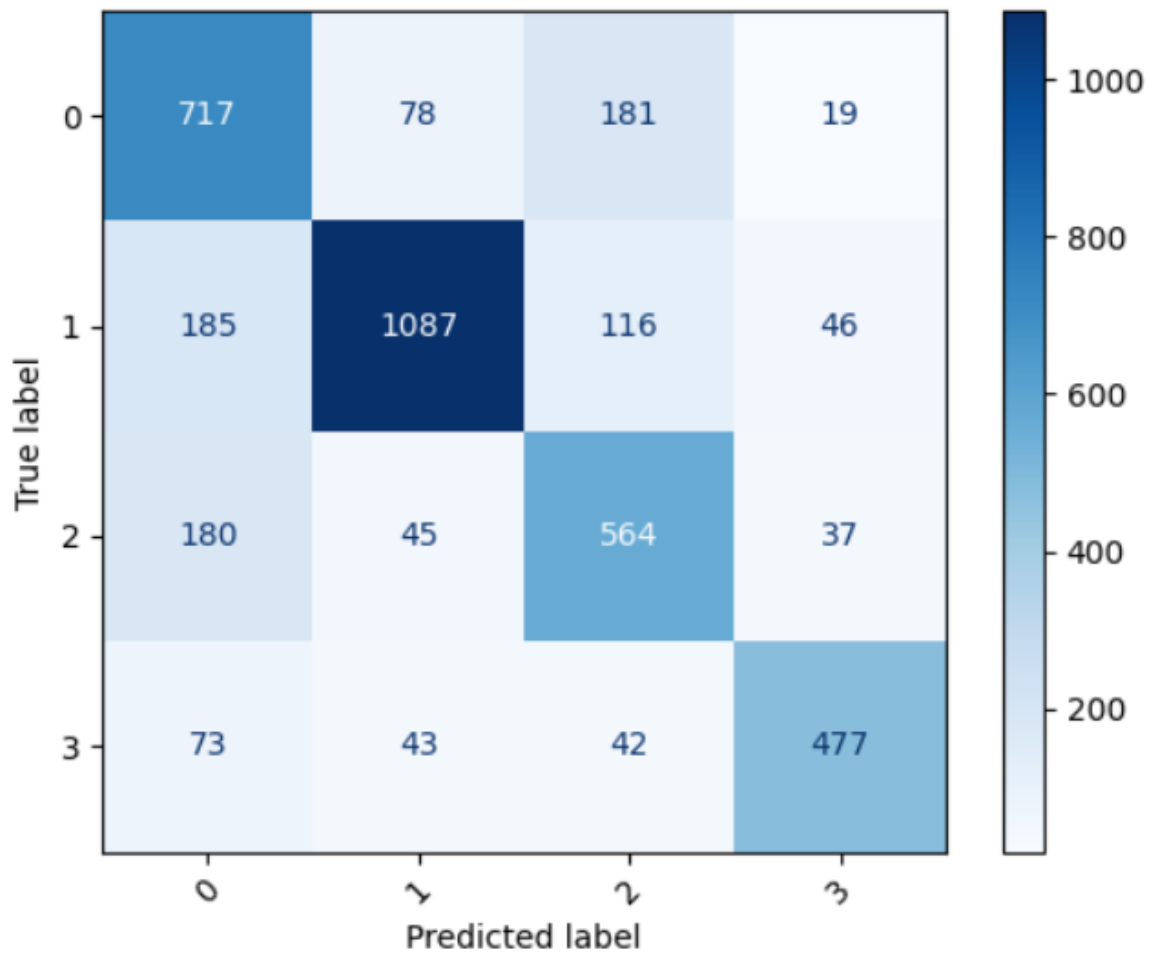

Figure (20) four emotions confusion matrix

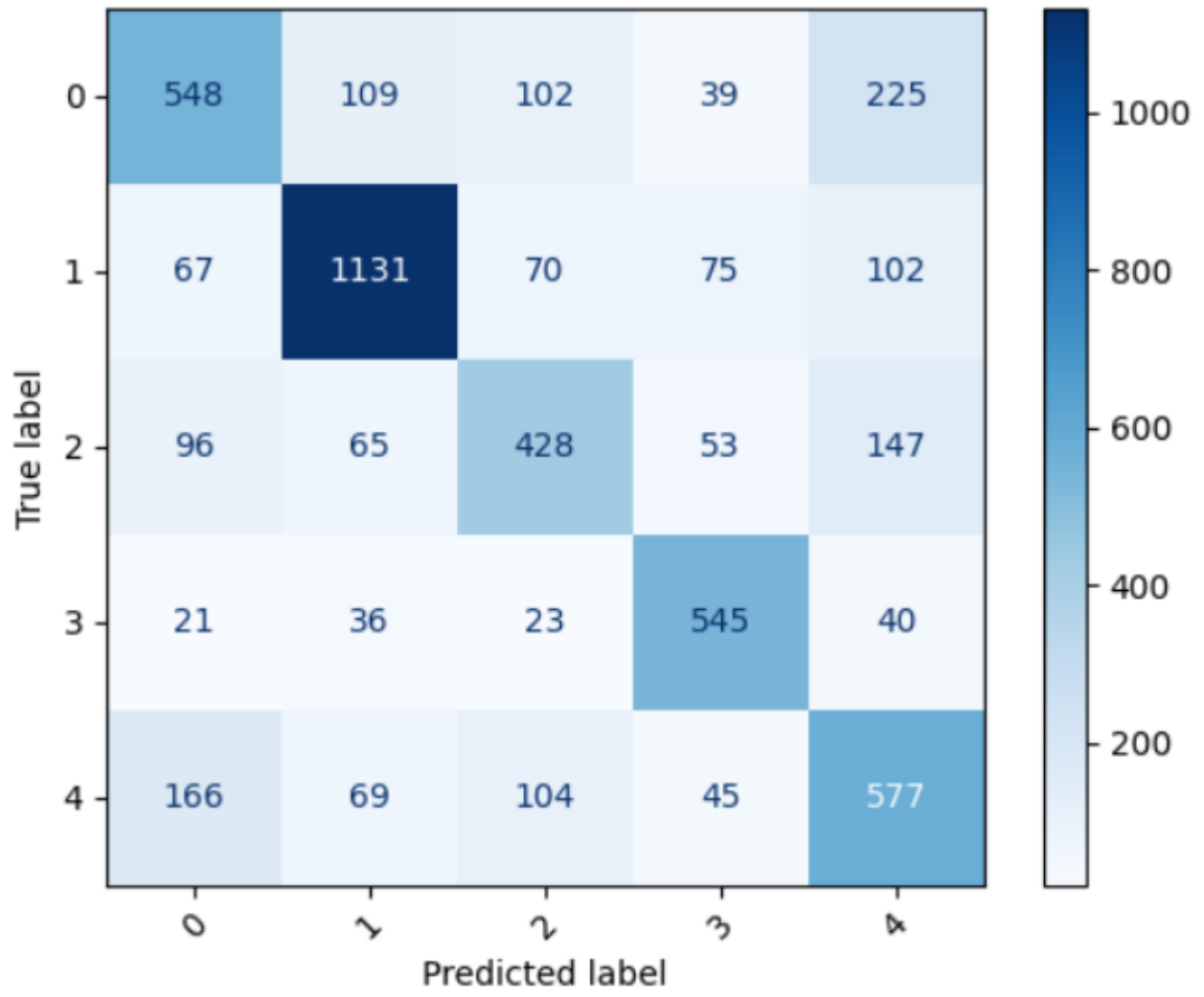

Figure (21) five emotions confusion matrix

| class | precision | recall | f1-score |
|---|---|---|---|
| neutral | 61% | 54% | 57% |
| Happy | 80% | 78% | 79% |
| Angry | 59% | 54% | 56% |
| surprise | 72% | 82% | 77% |
| sad | 53% | 60% | 56% |

Figure (22) classification report for the final used model (5 emotions)

Given the complexity of the task, low-quality dataset, and emotions are indeed overlapping and varying from each person, these results are considered very sufficient. During testing, we added a percentage prediction for the two highest emotions and by doing this, we improved the predictions and gave a better estimate of how people usually have mixed feelings.

*5.5 GUI*

To Visualize our results, we developed "IdentiFace" which is a Pyside based desktop application using Python. The GUI mainly consists of:

- A welcome landing window
- An offline window: where you can upload an image and perform the required classification/prediction
- An online window: where you can open your laptop camera and perform real-time detection.

Note that the recognizer demands a high quality images so to overcome this, we only added the recognizer to the offline window

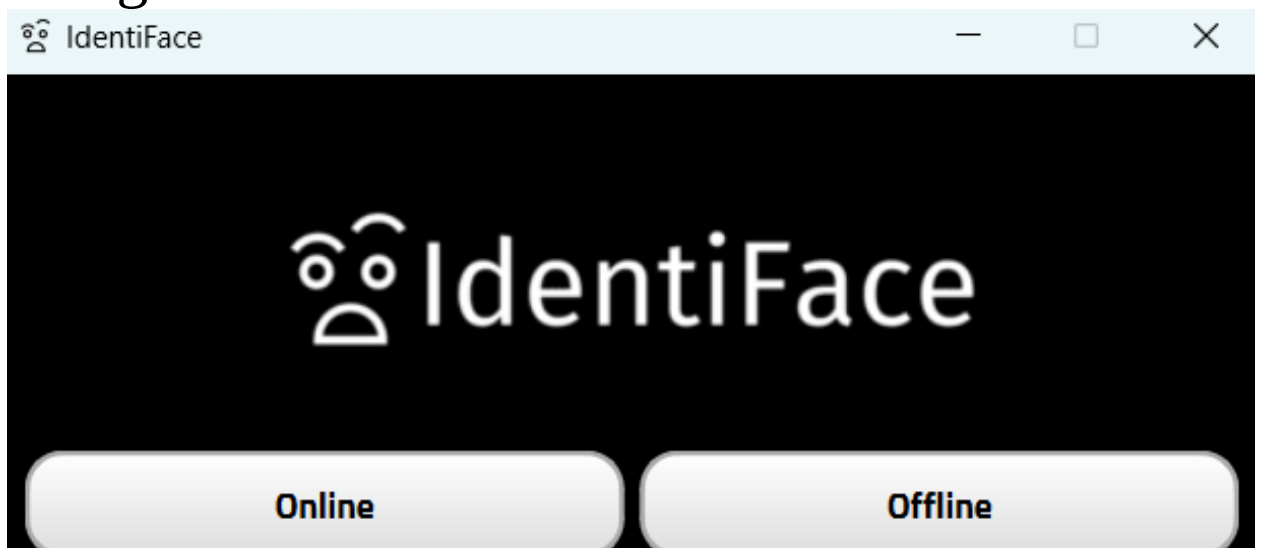

[6] Welcome window

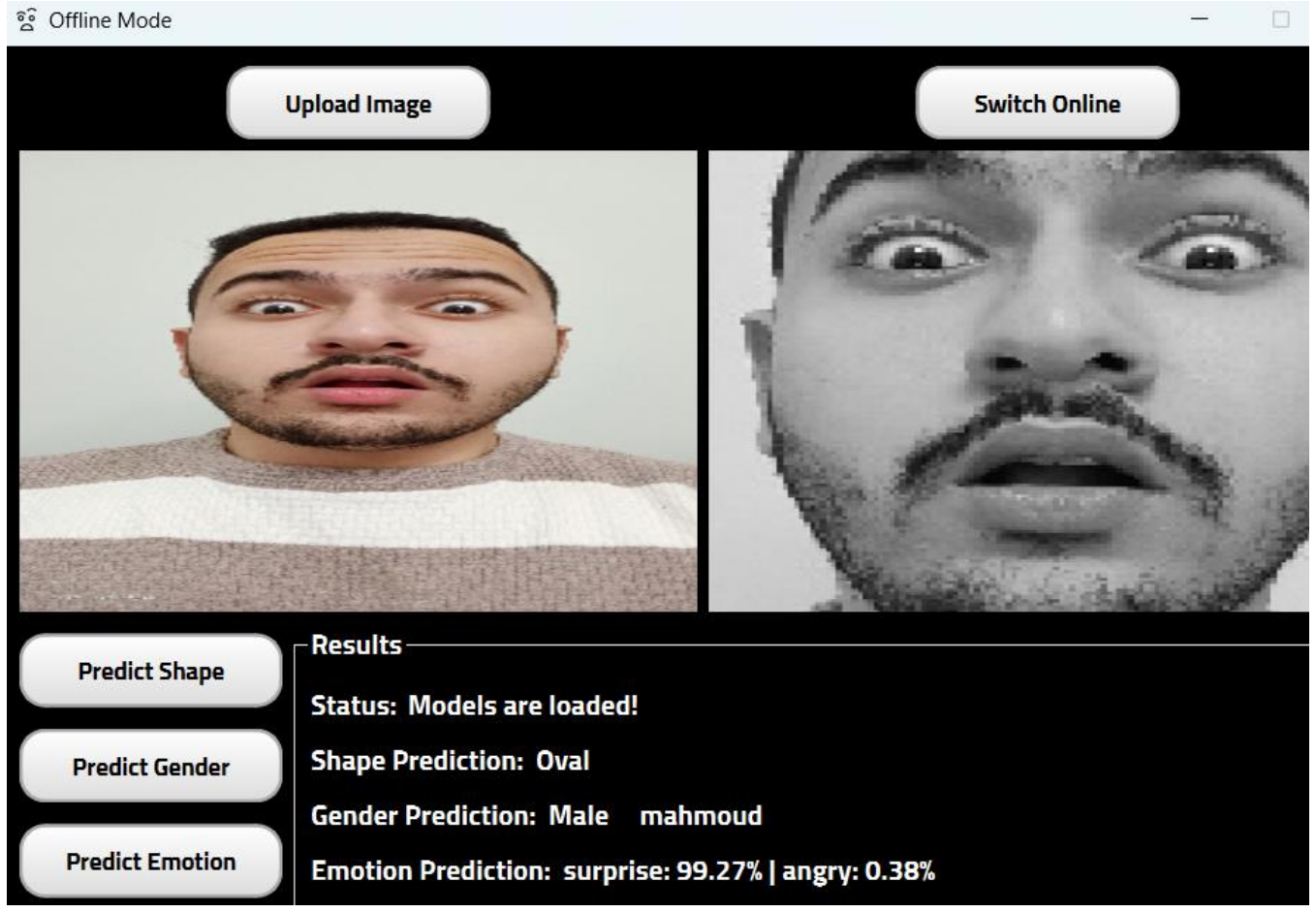

[7] Offline mode

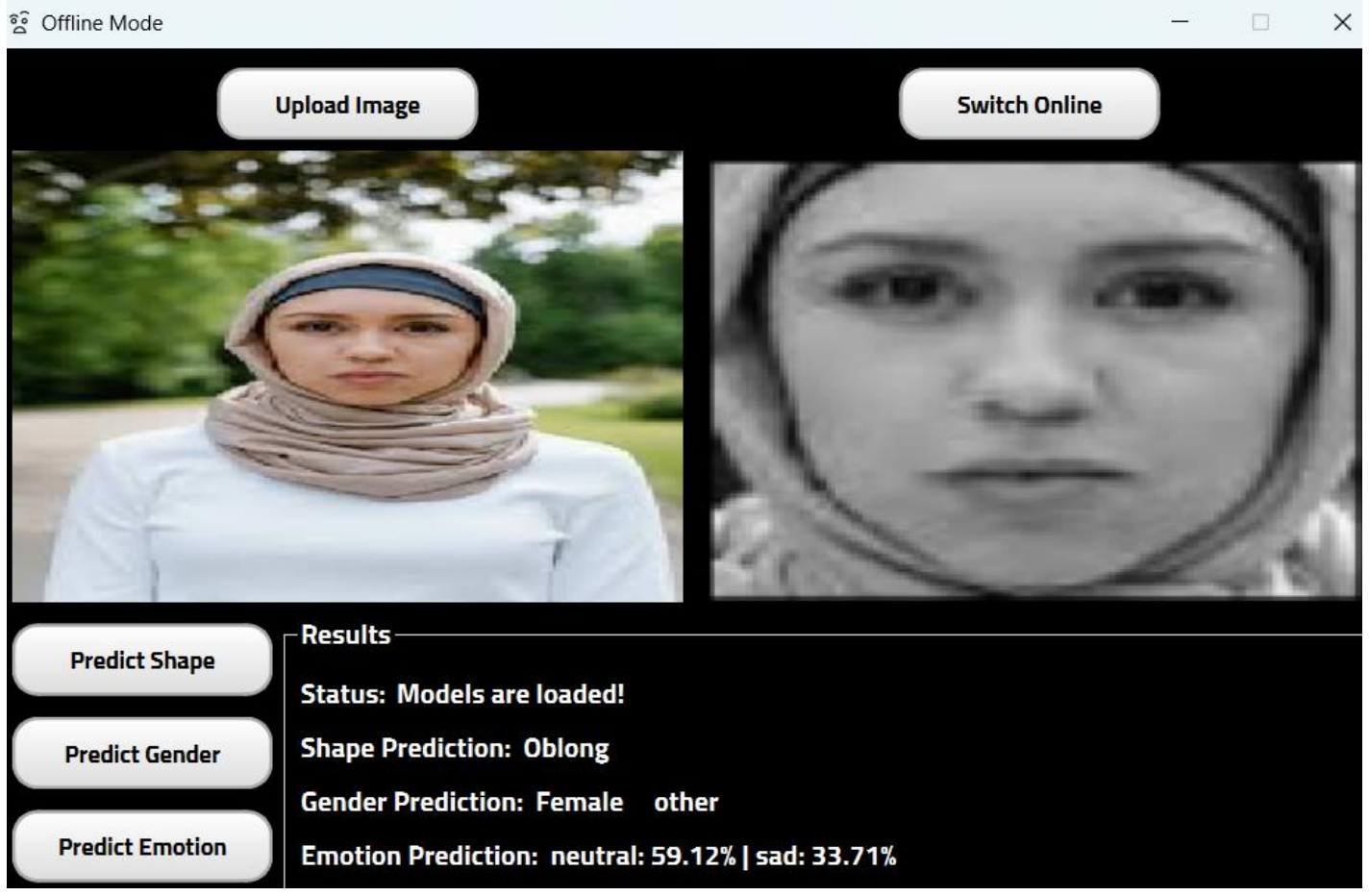

[8] Offline mode 2

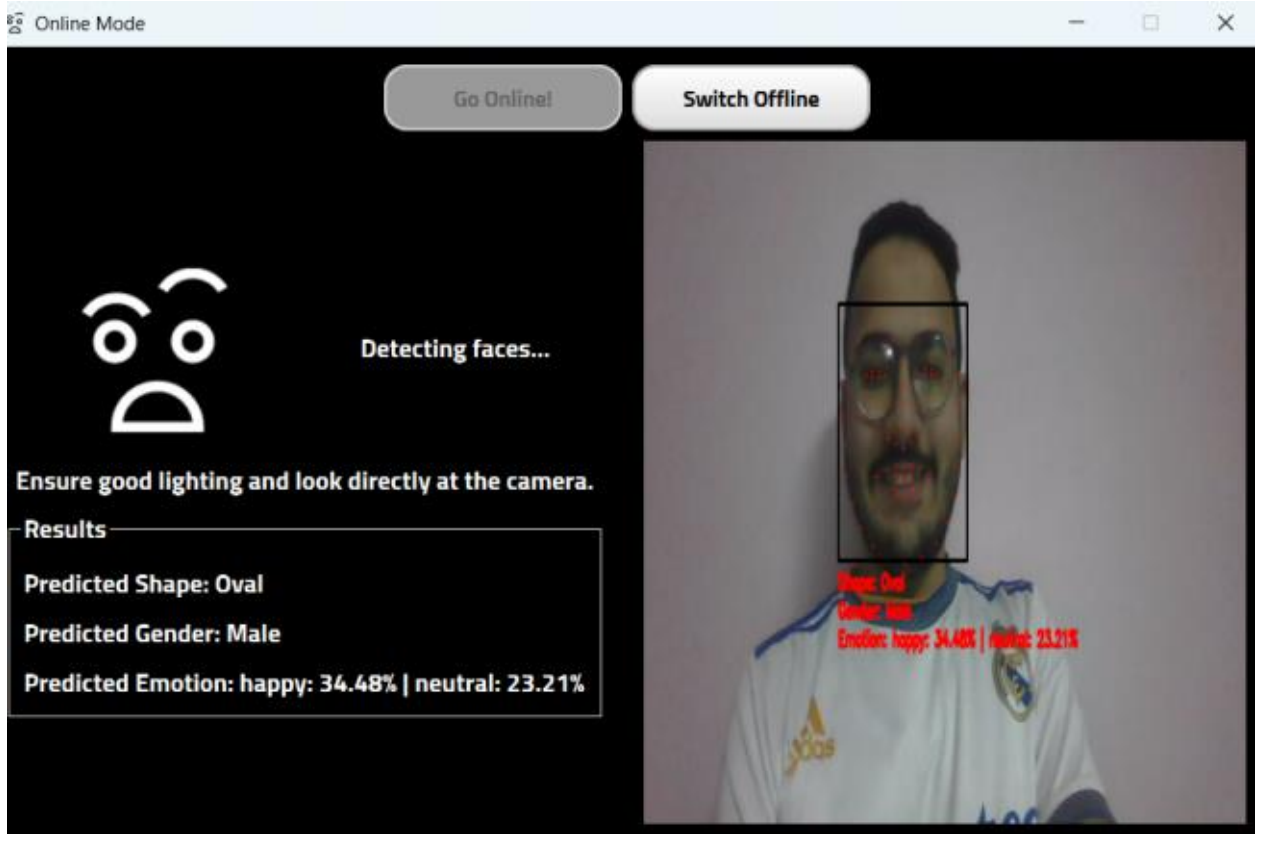

[9] Online mode

## Conclusion

After taking many approaches and trying different techniques, collecting our dataset for each task, and using other datasets in face recognition, gender classification, face shape detection, and emotion recognition, we decided to use the VGGNet model as it showed the highest results in all the addressed tasks using the following datasets: color FERET[1], a public dataset [2] for gender classification, the celebrity face shape [3] for face shape detection, and FER-2013 dataset [4] for emotion recognition. We also combined all the best-performing output models into one system called IdentiFace, a multimodal facial biometric system that combines facial recognition with gender, face shape, and emotion. Finally, we have a fully operational facial biometric system based on VGGNet architecture that can identify people, genders, face shapes, and emotions in real-time and offline.

**Acknowledgements**

We would like to thank everyone who helped in this project, especially Laila Abbas the TA. We would also like to thank our colleagues who have participated in the data collection process and everyone on the biometric course.